\pdfoutput=1

\documentclass[11pt]{article}

\usepackage[preprint]{acl}

\usepackage{times}
\usepackage{mdframed}
\usepackage{latexsym}

\usepackage[T1]{fontenc}

\usepackage[utf8]{inputenc}

\usepackage{microtype}

\usepackage{inconsolata}

\usepackage[utf8]{inputenc}
\usepackage[T1]{fontenc}
\usepackage{graphicx}
\usepackage{makecell}
\usepackage{booktabs}
\usepackage{multirow}
\usepackage{color}
\usepackage{longtable}
\usepackage{amsmath}
\usepackage{booktabs}
\usepackage{makecell}
\usepackage{float}
\usepackage{subfigure}
\usepackage{multirow}
\usepackage{wrapfig}
\usepackage{bbding}
\usepackage{arydshln}
\usepackage{lipsum}
\usepackage{mdframed}

\usepackage{xcolor} 
\usepackage{colortbl} 
\usepackage{fontawesome}
\usepackage{amsmath, amssymb} 
\usepackage{wrapfig}
\usepackage{caption}
\title{RefCritic: Training Long Chain-of-Thought Critic Models \\ with Refinement Feedback}

\vspace{5px}
\author{
\hspace{-2px}
Qiaoyu Tang${}^{1,3}$\thanks{~ Equal contribution},
Hao Xiang${}^{1,3}$\footnotemark[1],
Le Yu${}^{2}$,
Bowen Yu${}^{2}$,
Hongyu Lin${}^{1}$,
Yaojie Lu${}^{1}$,
\\
\textbf{Xianpei Han}${}^{1}$,
\textbf{Le Sun}${}^{1}$,
\textbf{Junyang Lin}${}^{2}$ 
\vspace{5px}
\\
$^{\rm 1}$Chinese Information Processing Laboratory, Institute of Software, \\Chinese Academy of Sciences \\
$^{\rm 2}$Alibaba Group  \hspace{5px} $^{\rm 3}$University of Chinese Academy of Sciences \vspace{5px}  \\
\hspace{1px}\{tangqiaoyu2020, xianghao2022, hongyu, luyaojie, xianpei, sunle\}@iscas.ac.cn \\ \hspace{1px}\{chuanyi.yl, yubowen.ybw, junyang.ljy\}@alibaba-inc.com
}

\begin{document}
\maketitle
\begin{abstract}
With the rapid advancement of Large Language Models (LLMs), developing effective critic modules for precise guidance has become crucial yet challenging.
In this paper, we initially demonstrate that supervised fine-tuning for building critic modules (which is widely adopted in current solutions) fails to genuinely enhance models' critique abilities, producing superficial critiques with insufficient reflections and verifications.
To unlock the unprecedented critique capabilities, we propose RefCritic, a long-chain-of-thought critic module based on reinforcement learning with dual rule-based rewards: (1) instance-level correctness of solution judgments and (2) refinement accuracies of the policy model based on critiques, aiming to generate high-quality evaluations with actionable feedback that effectively guides model refinement.
We evaluate RefCritic on Qwen2.5-14B-Instruct and DeepSeek-R1-Distill-Qwen-14B across five benchmarks. On critique and refinement settings, RefCritic demonstrates consistent advantages across all benchmarks, e.g., 6.8\% and 7.2\% gains on AIME25 for the respective base models.
Notably, under majority voting, policy models filtered by RefCritic show superior scaling with increased voting numbers.
Moreover, despite training on solution-level supervision, RefCritic outperforms step-level supervised approaches on ProcessBench, a benchmark to identify erroneous steps in mathematical reasoning.
\end{abstract}

\section{Introduction}
In recent years, Large Language Models (LLMs) have demonstrated remarkable capabilities in executing complex reasoning tasks such as mathematical problem-solving and code generation~\cite{yang2025qwen3technicalreport,hui2024qwen25codertechnicalreport}. As these models continue to evolve, their reasoning processes have grown increasingly sophisticated, encompassing multiple elaborate steps and diverse pathways~\citep{guo2025deepseek,qwq32b}. This progression introduces a critical challenge: reasoning processes are becoming substantially difficult for humans to supervise, making errors within these intricate chains harder to identify and rectify. The escalating complexity of LLM-generated solutions necessitates more effective analytical frameworks to evaluate and enhance reasoning quality, extending beyond the constraints of human supervisory oversight.

Developing LLM critics has become a promising direction for evaluating complex reasoning tasks~\citep{liu2025inference, zhang2024generative, mahan2024generative, wang2023math, ankner2024critique}, functioning as specialized modules to analyze reasoning processes and identify errors. Ideally, LLM critics are expected to comprehensively analyze content generated by the policy model, delivering targeted critiques that identify logical inconsistencies or factual errors and improve the refinement quality of the policy model. However, contemporary approaches exhibit two critical limitations. Firstly, they frequently produce superficial evaluations characterized by insufficient analytical depth~\citep{zheng2025criticcot,tang2025enablingscalableoversightselfevolving} and typically necessitate granular step-level annotations of the solution for optimization~\citep{yang2025deepcriticdeliberatecritiquelarge}. Secondly, current implementations mainly focus on metrics of critic performance while overlooking the practical utility of critiques in enhancing policy model refinement.

\begin{figure*}[t!]
\centering 
\includegraphics[width=\textwidth]{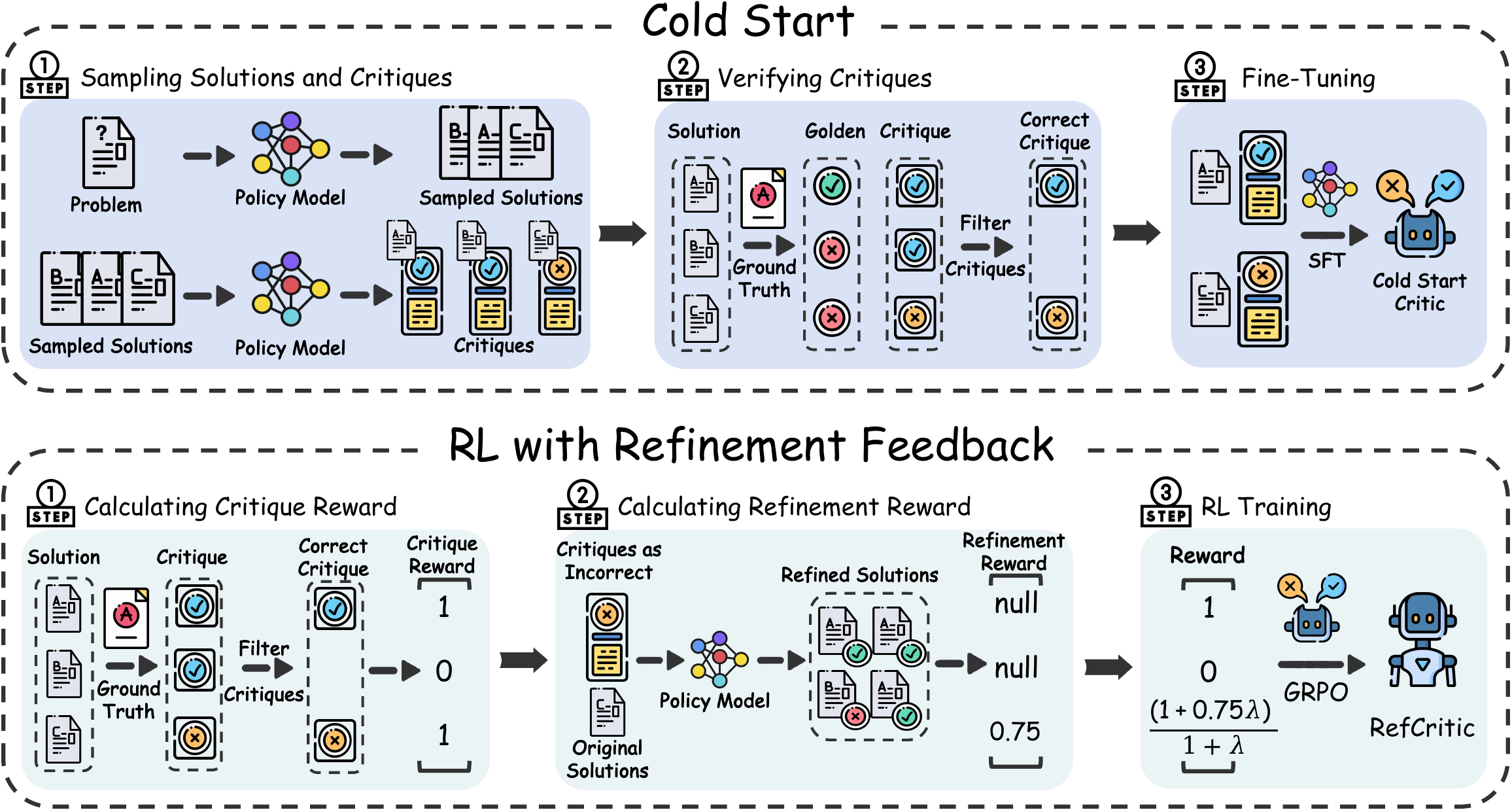}
\caption{The Critic model with Refinement Feedback RefCritic framework consists of two steps: (1) cold-start via rejective sampling fine-tuning, (2) rule-based reinforcement learning with refinement feedback. With this two-stage optimization, RefCritic generates in-depth critiques that achieve superior critic performance and effectively guide policy model refinement through actionable feedback.}

\label{fig:method}
\end{figure*}

In this paper, we propose RefCritic, a long chain-of-thought critic model with refinement feedback to tackle the above limitations, which could generate in-depth critiques that not only achieve superior critic performance but also effectively guide policy model refinement through actionable feedback. This process begins with prompting open-source models (e.g., DeepSeek-R1-Distill-Qwen) to generate seed data containing three essential components: long CoT analysis, solution validity judgments, and refinement suggestions. After rigorous quality filtering based on judgment accuracy, about 10K valid samples are obtained, which are utilized to establish cold-start critic models via Supervised Fine-Tuning (SFT). We then conduct preliminary assessments to verify whether SFT is sufficient for producing comprehensive critiques. Results reveal that SFT alone struggles to produce in-depth critiques despite generating lengthy CoT content, as models frequently exhibit misleading analytical patterns where correct judgments emerge from flawed reasoning processes (a persistent issue observed in existing LLM critics), resulting in unreliable performance evaluations. Furthermore, the absence of explicit policy model interaction during SFT leads to critiques that inadequately prioritize feasible guidance for effective refinement. The above observations highlight the difficulty of SFT in producing critiques with both accurate evaluations and practical feedback for refinement.

To further enhance the critic reliability and establish a causal connection between critique quality and policy refinement outcomes, we implement a dual-reward reinforcement learning framework based on cold-start models for finalizing RefCritic. The first reward signal stems from the instance-level binary accuracy metric (0/1 values) for evaluating the critic models in solution judgment capability. The second reward quantifies policy model improvement through accuracy gains after incorporating refinement suggestions. Critically, the second reward establishes an explicit feedback loop where the quality of effective critique is operationally defined by its capacity to drive measurable enhancements to the policy model. This dual-reward design ensures that high-reward critiques are those that not only accurately identify solution flaws but also provide actionable guidance leading to verifiable performance gains. 

We validate the effectiveness of RefCritic on multiple challenging mathematical datasets: AIME24, AIME25, and Olympiad~\citep{he2024olympiadbench}. In the refinement after critique setting, feedback generated by RefCritic based on Qwen2.5-14B and DeepSeek-R1-Distill-Qwen-14B consistently enhances the corresponding policy models' performance, with improvements of 6.8\% and 7.2\% on AIME25, and 9.9\% and 2.6\% on Olympiad, respectively. In the majority vote with critique setting, RefCritic demonstrates increasingly significant performance gains as the sampling count increases. With 64 samples, RefCritic achieves an average improvement of 3.6 percentage points on AIME25 compared to scenarios without critique, consistently outperforming other critique baselines. Moreover, RefCritic effectively enhances majority vote performance even when applied to more powerful models, like QwQ and DeepSeek-Distill-Qwen-32B. These performance improvements indicate that our dual reward mechanism successfully aligns critique generation with both evaluation accuracy and refinement utility, enabling critic models to produce not only precise solution assessments but also actionable feedback that effectively guides refinement processes. Furthermore, it is worth noting that RefCritic generalizes effectively to step-level critique tasks without requiring any step-level labels during training, achieving remarkable performance on ProcessBench~\citep{zheng2024processbench}.

\section{Related Work}

\paragraph{Test-time Scaling}
Test-time scaling techniques have emerged as a powerful approach to enhance LLM reasoning capabilities through increased computational resources during inference~\citep{charniak2005coarse, snell2024scaling,wu2024empirical, yao2023tree, chen2024boosting, jaech2024openai,guo2025deepseek,qwq32b}. The effectiveness of these approaches can be improved by judgment or verification mechanisms. Besides traditional process reward models (PRMs) that directly predict numerical correctness scores for solution steps \citep{uesato2022solving, lightman2023let,zheng2025criticcot, zheng2023judging, chen2025spcevolvingselfplaycritic, prmlessons}, recent methods frame judgement as language generation tasks that offer greater interpretability and scalability \citep{liu2025inference, zhang2024generative, mahan2024generative, wang2023math, ankner2024critique}. Among them, a promising approach is LLM Critics, which uses LLMs as critic models to verify solutions~\citep{mcaleese2024llm,zheng2025criticcot,yang2025deepcriticdeliberatecritiquelarge}.

\paragraph{Critique for Math}
The judgment ability of LLMs has garnered significant research interest due to their potential to enhance mathematical reasoning through explicit error detection and correction guidance~\citep{lan2024criticeval, lin2024criticbench, zheng2024processbench}. Current approaches fall into two main categories: LLM-as-a-Critic~\citep{zheng2025criticcot, yang2025deepcriticdeliberatecritiquelarge} leverages off-the-shelf models through careful instruction design, and specialized critic models\citep{mcaleese2024llm, lan2024criticeval, shi2025heimdalltesttimescalinggenerative} that employ fine-tuning or reinforcement learning to enhance judgment ability. While recent research emphasizes that critique effectiveness should be validated through correction outcomes \citep{tang2025realcriticeffectivenessdrivenevaluationlanguage, zheng2024processbench}, most existing critic models focus exclusively on critique and ignore the future benefit it can bring to refinement. Our work addresses these challenges through RefCritic, a novel framework that leverages reasoning models' critique abilities and incorporates refinement performance as a direct reward signal during critic training.
The most closely related concurrent works are ThinkPRM \citep{khalifa2025processrewardmodelsthink} and DeepCritic \citep{yang2025deepcriticdeliberatecritiquelarge}. The former uses SFT to enhance the capabilities of reasoning models, while the latter improves the critique performance of instruction models on individual steps through complex fine-tuning and simple RL. However, they fail to recognize how critiques serve as valuable feedback mechanisms for policy model refinement.

\section{SFT is Insufficient for Deep Critiques}
\label{pre_ex}
To better understand the challenges in developing effective critic models, we first examined a straightforward approach widely adopted in previous research: supervised fine-tuning with rejection sampling. This approach has demonstrated success in improving judgment capabilities of critics in several studies~\citep{tang2025enablingscalableoversightselfevolving,zheng2025criticcot,khalifa2025processrewardmodelsthink}.

Specifically, we employed Qwen2.5-14B-Instruct/DeepSeek-Distill-Qwen-14B as our policy model and DeepSeek-Distill-Qwen-32B as the critic model\footnote{For Qwen2.5-14B-Instruct, we provide an empty thinking process and only use the content after "</think>".}. Employing rejection sampling, we collected approximately 10K critique training samples from a subset of NuminaMath. Each training sample comprised a problem statement, model response, chain-of-thought reasoning, judgment on solution correctness, and refinement suggestions. We subsequently fine-tuned the policy models on these datasets to develop critique capabilities. To evaluate the effectiveness of these fine-tuned critic models, we tested them on responses collected from AIME25, assessing their ability to identify errors and provide feedback that could meaningfully improve policy model performance.

\begin{table}[ht]
\centering
    \resizebox{0.4\textwidth}{!}{
    \begin{tabular}{lcc}
    \hline
    \multirow{1}{*}{\makecell[c]{Method}}  & \makecell[c]{Critique\\Accuracy} & \makecell[c]{Pass@1 after\\Refinement} \\ 
    \hline
    \multicolumn{3}{l}{\textcolor{lightgray!99}{\textit{Qwen2.5-14B as Base Model}}} \\
    Pass@1       & -        & 14.4   \\
    Self-Critique    & 51.8     & 14.5   \\
    SFT           & 80.6     & 15.0   \\
    \noalign{\vskip 0.11cm}
    \hdashline
    \noalign{\vskip 0.11cm}
    \multicolumn{3}{l}{\textcolor{lightgray!99}{\textit{R1-Qwen-14B as Base Model}}} \\
    Pass@1          & -        & 49.2   \\
    Self-Critique & 71.5     & 52.1   \\
    SFT           & 78.9     & 51.4   \\ 
    \hline
    \end{tabular}
    }
    \caption{
        Preliminary experiment on AIME25 to verify whether SFT can emerge deep critic. We can see that although the SFT model achieves strong performance in critique evaluation, incorporating its feedback into the policy model yields only marginal performance gains. R1-Qwen represents DeepSeek-Distill-Qwen.
    }
\label{tab:preliminary}
\vspace{-10px}
\end{table}
Our experiments revealed a significant disparity between the critic model capabilities and their practical utility. As shown in Table~\ref{tab:preliminary}, whether it is a Qwen-based critic or DeepSeek-Distill-Qwen-based critic, SFT-trained models significantly outperformed self-critique approaches in critique accuracy metrics. However, when these critiques were used to refine policy model outputs, the resulting performance improvements were minimal, sometimes even inferior to those achieved through self-critique methods. This counterintuitive result suggests that conventional evaluation metrics for critics may not align with their actual utility in improving model performance.

Further analysis of the SFT model outputs revealed two critical limitations. First, critics often arrive at correct judgments through flawed or superficial reasoning processes, creating a false impression of reliability despite inconsistent analytical quality. This problem is particularly evident in Qwen-based models, as the critique length after SFT showed no significant increase, with an average length of less than 500 tokens.  Second, the feedback typically identified error locations but lacked specific, actionable guidance for improvement. Critics frequently offered vague suggestions or restated problem requirements rather than providing concrete directions for correcting mathematical misconceptions or reasoning flaws. These findings directly support our hypothesis that conventional SFT approaches, while successful in training critics to make binary judgments, fail to develop models that can provide the actionable, improvement-oriented feedback necessary for effective solution refinement.

\section{RefCritic}
We propose RefCritic, a novel approach for developing effective critic modules that provide actionable feedback for mathematical reasoning tasks. As illustrated in Figure~\ref{fig:method}, RefCritic employs a two-stage methodology. First, we develop a cold-start critic model via supervised fine-tuning that activates the model's reasoning judgment capabilities and enables structured output generation. Second, we introduce a rule-based reinforcement learning framework with dual rewards optimizing critics for both solution-level correctness and refinement effectiveness, measured by concrete improvements in policy model performance. This dual-reward mechanism ensures our critic models not only accurately evaluate solutions but also provide guidance that leads to substantive improvements in reasoning capabilities.

When faced with complex tasks such as critique generation, LLMs often exhibit problematic behaviors, including instruction unfollowing~\citep{he2025largelanguagemodelsdetect} and answer leakage~\citep{yang2025deepcriticdeliberatecritiquelarge}. To address these challenges, we implement SFT with rejection sampling to standardize model outputs. Following our preliminary experimental setup, we leverage a more powerful model to generate initial critic responses, then systematically filter out responses containing erroneous judgments, instruction violations, or solution leakage risks through rule-based screening. This curated dataset serves as the foundation for our LLM training, ensuring the resulting critic model adheres to desired output formats while maintaining evaluation integrity.

Despite the effectiveness of the supervised fine-tuning approach in producing format-compliant critics, the preliminary experiments revealed fundamental shortcomings that limit practical utility. As previously demonstrated, SFT models exhibit a misleading combination of accurate judgments built upon superficial reasoning, alongside feedback that identifies errors without providing actionable remediation strategies. Rather than reiterating these limitations, we recognize them as symptomatic of a deeper issue: conventional training methods optimize for solution classification accuracy rather than refinement capability. This insight motivates us to shift from purely supervised learning toward a reinforcement learning framework that explicitly rewards critics not just for evaluation correctness, but for generating feedback that demonstrably improves subsequent solutions.

To address these limitations, we introduce a dual-reward reinforcement learning framework that optimizes both judgment accuracy and refinement effectiveness. Our approach evaluates critics based on two key metrics: (1) their ability to correctly classify solutions as right or wrong, and (2) the tangible improvement their feedback produces when used by the policy model to revise incorrect solutions. This framework ensures our critics develop both strong evaluation capabilities and the ability to generate constructive feedback that leads to measurable improvements in reasoning outcomes.

Formally, let $C_\theta$ represent our critic model with parameters $\theta$, and $P_\phi$ denote the policy model with parameters $\phi$ that generates mathematical solutions. For a problem $x$ with ground truth answer $a$ from dataset $\mathcal{D}$, the policy model produces an initial solution $y_0 \sim P_\phi(y|x)$. The correctness of this solution is determined by $c = \mathbb{I}[y_0 = a] \in \{0, 1\}$, where $\mathbb{I}[\cdot]$ is the math equal function.
The critic model then performs an evaluation $C_\theta(x, y_0) \rightarrow (z, \hat{c}, f)$, where $z$ represents an extensive reasoning process, $\hat{c} \in \{0, 1\}$ denotes the predicted correctness judgment, and $f$ provides actionable refinement suggestions. Based on these critiques, the policy model generates $m$ refined solutions $\{y_i\}_{i=1}^m$ where $y_i \sim P_\phi(y|x, y_0, f)$ for $i \in \{1, 2, \ldots, m\}$.
Our reinforcement learning objective maximizes:
\begin{equation*}
    \begin{aligned}
    \mathcal{J}(\theta) &= \mathbb{E}[R_j(c, \hat{c}) + \lambda R_r(c, \hat{c}, a, \{y_i\}_{i=1}^m)]
    \end{aligned}
    \label{eq:GRPO-obj}
\end{equation*}
where $R_j$ represents the binary judgment score:
\begin{equation*}
R_j(c, \hat{c})=\left\{
\begin{array}{rcl}
1     &      & \text{if } c = \hat{c},\\
0     &      & \text{otherwise}\\
\end{array} \right.
\label{eq:reward_judgment}
\end{equation*}
and $R_r$ denotes the refinement score, which is non-zero only when the critic correctly identifies an incorrect solution ($c = 0$ and $\hat{c} = 0$):
\begin{equation*}
\resizebox{0.49\textwidth}{!}{$
R_r(c, \hat{c}, a, \{y_i\}_{i=1}^m)=\left\{
\begin{array}{lcl}
\frac{1}{m} \sum\limits_{i=1}^{m} \mathbb{I}[y_i = a]     &      & \text{if } c = 0 \text{ and } c = \hat{c} \\
0       &      & \text{otherwise}\\
\end{array} \right.
$}
\label{eq:reward_refinement}
\end{equation*}
$\lambda$ is a hyperparameter that balances the importance between judgment accuracy and refinement effectiveness. A higher $\lambda$ value places greater emphasis on the critic's ability to provide actionable feedback that leads to correct solutions, while a lower value prioritizes accurate solution classification.

In summary, our RefCritic framework alleviates the key limitations of existing critic models through this dual-reward reinforcement learning approach. By explicitly optimizing for both judgment accuracy and refinement effectiveness, we develop critics that not only accurately evaluate mathematical solutions but also provide actionable feedback that leads to concrete improvements in reasoning outcomes.
\section{Experiments}

\begin{table*}[!t]
    \centering
    \resizebox{\textwidth}{!}{
        \begin{tabular}{l|ccc|ccc|ccc}
        \toprule
        \multirow{2}[1]{*}{Model} & \multicolumn{3}{c}{AIME25} & \multicolumn{3}{c}{AIME24} & \multicolumn{3}{c}{Olympiad} \\
        \cmidrule(lr){2-4} \cmidrule(lr){5-7} \cmidrule(lr){8-10}
              & $Pass_r@1$ & $Maj_c@8$ & $Maj_c@64$ & $Pass_r@1$ & $Maj_c@8$ & $Maj_c@64$ & $Pass_r@1$ & $Maj_c@8$ & $Maj_c@16$ \\
            \midrule
            Qwen-14B Majority Vote & 14.4  & 19.2  & 23.3  & 13.7  & 16.5  & 16.6  & 45.8  & 52.2  & 53.6  \\
            Qwen-14B as Critic & 14.5  & 19.1  & 22.7  & 13.7  & 18.5  & 21.2  & 45.8  & 52.3  & 54.0  \\
            \noalign{\vskip 0.11cm}
            \hdashline
            \noalign{\vskip 0.11cm}
            \multicolumn{10}{l} 
            {\textcolor{black}{\textit{RefCritic-Qwen-14B(Ours)}}} \\
            \quad $SFT$ & 15.0  & 19.3  & 23.4  & 15.2  & 19.2  & 23.9  & 46.6  & 52.5  & 54.3  \\
            \quad $RL_{\lambda=0}$  & 18.5  & 20.8  & 22.4  & 19.1  & 20.5  & 23.8  & 51.4  & 55.4  & 57.4  \\
            \quad $RL_{\lambda=0} \xrightarrow{after} RL_{\lambda=1}$ & \textbf{21.2} & \textbf{21.5} & \textbf{24.4} & \textbf{23.0} & \textbf{21.4} & \textbf{26.6} & \textbf{55.7} & \textbf{57.3} & \textbf{59.2} \\
            \midrule
            R1-Qwen-14B Majority Vote & 49.1  & 61.6  & 62.0  & 67.6  & 78.7  & 80.1  & 77.7  & 82.7  & 83.3 \\
            R1-Qwen-14B as Critic & 50.0  & 60.6  & 62.9  & 70.5  & 79.3  & 82.4  & 78.8  & 82.7  & 83.3 \\
            \noalign{\vskip 0.11cm}
            \hdashline
            \noalign{\vskip 0.11cm}
            \multicolumn{10}{l}{\textcolor{black}{\textit{RefCritic-R1-14B(Ours)}}} \\
            \quad $SFT$  & 51.3  & 61.6  & 62.8  & 71.4  & 79.4  & \textbf{83.1}  & 78.7  & 83.0    & 84.4 \\
            \quad $RL_{\lambda=0}$  & 55.1 & 64.2 & 67.1	& \textbf{73.5} & \textbf{80.4} & 82.8 & \textbf{80.4} & 83.8 & 84.5 \\
            \quad $RL_{\lambda=0}$ $\xrightarrow{after}$ $RL_{\lambda=1}$ & \textbf{56.3}  & \textbf{65.2}  & \textbf{68.1}  & 72.8  & \textbf{80.4}  & 82.5  & 80.3  & \textbf{83.9}  & \textbf{84.7} \\
            \bottomrule
        \end{tabular}
  }
    \caption{Performance comparison of different approaches on AIME24/25 and Olympiad. $Pass_r$ indicates the performance after one round of critique and refinement. $Maj_c$ indicates the majority vote performance after using the critic filtering solutions. $RL_{\lambda=1}$ indicate RL with Refinement Feedback. Considering the cost of sampling refinements, we initially set $\lambda=0$ to achieve rapid improvement in critic performance.}
        \label{tab:main_results}
\end{table*}

\subsection{Experimental Setup}
\label{setting}
\paragraph{Models} For our implementation of RefCritic, we utilize Qwen2.5-14B-Instruct~\citep{yang2025qwen3technicalreport} and DeepSeek-R1-Distill-Qwen-14B~\citep{guo2025deepseek} as the backbone. 
In our framework, these models perform two distinct functions: first, as policy models that generate solutions for mathematical problems; and second, as the foundation models to develop our critic models.

\paragraph{Data Construction}
We construct our training dataset by filtering approximately 120k high-quality mathematical problems from the 900k problems in NuminaMath-1.5~\citep{numina_math_datasets}. Our filtering pipeline includes: (1) deduplication through exact string matching and semantic similarity using gte-multilingual-base embeddings (removing pairs with cosine similarity > 0.95); (2) problem filtering using Qwen2.5-72B-Instruct to remove unsolvable, proof, and multiple-choice problems; (3) difficulty balancing by sampling eight solutions and excluding problems where all attempts succeed or fail. For critic training, we sample 8 responses per problem and retain at most two responses per problem (one correct, one incorrect) to ensure balanced training data. Detailed process can be found in Appendix \ref{data_filter}.

\paragraph{Benchmarks}
We evaluate the performance of RefCritic on challenging mathematical benchmarks, including AIME 2024/2025 (American Invitational Mathematics Examination problems), and OlympiadBench~\citep{he2024olympiadbench} (a collection of mathematical Olympiad problems).
Since RefCritic was trained only on math problems, we conduct out-of-distribution (OOD) generalization tests on the code generation task LiveCodeBench~\citep{jain2024livecodebench} and the science QA benchmark GPQA-Diamond~\citep{rein2024gpqa}.
Furthermore, to evaluate RefCritic’s capability for fine-grained error localization, we leverage ProcessBench~\citep{zheng2024processbench} to assess its ability to accurately identify the specific step where an error occurs.

\paragraph{Evaluation Details}
For evaluation, both policy and critic models use a sampling strategy with temperature=0.6 and $top\_p$=0.95. For AIME24/25, we pre-sample 128 responses as the response pool for subsequent performance calculations. For OlympiadBench/GPQA/LiveCodeBench, we sample only 32 responses due to its larger scale. During evaluation, we randomly select responses from the response pool for metric calculation and report the average results over 1000 trials.
We adopt multiple evaluation settings:

\textit{Majority Vote with Critique}: The critic model first evaluates each of the N sampled solutions and filters out those judged as incorrect. We then apply majority vote to the remaining solutions to select the final answer, denoted as $Maj_c@N$. As a baseline, we also report standard majority vote accuracy without critique filtering.

\textit{Refinement after Critique}: The policy model generates an initial solution, which is then critiqued by the critic model. If judged incorrect, the policy model refines the solution based on the critique feedback. We report pass@1 accuracy of the final refined answer, denoted as $Pass_r@1$.

\textit{Process Critique Evaluation}: For process-level evaluation, since our critic models were trained to output natural language critiques rather than explicit step indices, we use Qwen2.5-14B-Instruct to identify the step index the critic judges incorrect.\footnote{We only provide the solution and critique, without the corresponding problem. Every critic we evaluated would go through this process.} Following ProcessBench, we report the F1 score, which is the harmonic mean of precision for correct and incorrect solutions.

\paragraph{Training Details}
In the SFT stage, we train the critic models with a learning rate of 7e-6 and a batch size of 512 for three epochs. For the RL stage, we employ the GRPO algorithm~\citep{shao2024deepseekmathpushinglimitsmathematical} to enhance critic performance. We sample 8 critics for each input, each rollout comprising 128 inputs, and conduct on-policy training with a learning rate of 1e-6. We set the maximum sequence length to 8K and 16K tokens for Qwen2.5-Instruct and DeepSeek-R1-Distill-Qwen, respectively. For refinement feedback, we use policy models to generate 8 refinements for each critic. Considering the cost of sampling refinements, we initially set $\lambda$=0 to achieve rapid improvement in critic performance for 600 steps, where no refinement is generated. We subsequently adjust to $\lambda$=1 to balance the trade-off between the two reward components and continue training for 300 steps.

\begin{table*}[!t]
    \centering
    \selectfont
    \resizebox{0.7\textwidth}{!}{
    \begin{tabular}{lccccc}
        \toprule
        Model & GSM8K & MATH  & Omni-Math & Olympiad & Avg. \\
        \midrule
        \multicolumn{6}{l}{\textcolor{lightgray!99}{\textit{PRM}}} \\
        Math-Shepherd-PRM-7B* & 47.9 & 29.5 & 24.8 & 23.8 & 31.5 \\
        RLHFlow-PRM-8B-DS* & 38.8 & 33.8 & 16.9 & 16.9 & 26.6 \\
        Qwen2.5-Math-PRM-7B* & 68.2 & 62.6 & 50.7 & 44.3 & 56.5 \\
        \noalign{\vskip 0.11cm}
        \hdashline
        \noalign{\vskip 0.11cm}
        \multicolumn{6}{l}{\textcolor{lightgray!99}{\textit{Prompt LLM as Critic}}} \\
        Qwen2.5-14B-Instruct & 61.7 & 52.6 & 41.3  & 43.1  & 49.7  \\
        Qwen2.5-72B-Instruct & 74.6 & 61.8 & 51.7 & 52.8 & 60.2  \\
        R1-Qwen-7B & 75.3  & 74.4  & 56.9  & 63.5  & 67.5  \\
        R1-Qwen-14B & 75.9  & 76.2  & 59.6  & 63.6  & 68.8  \\
        GPT-4o-0806* & 79.2 & 63.6 & 51.4 & 53.5 & 61.9 \\
        \noalign{\vskip 0.11cm}
        \hdashline
        \noalign{\vskip 0.11cm}
        \multicolumn{6}{l}{\textcolor{lightgray!99}{\textit{Baseline Critic}}} \\
        SCRIT-72B~\citep{tang2025enablingscalableoversightselfevolving} & 80.2  & 60.0  & 32.5  & 27.8  & 50.0  \\
        DeepCritic-7B~\citep{yang2025deepcriticdeliberatecritiquelarge} & 72.6  & 72.8  & 56.0  & 60.9 & 65.6  \\
        ThinkPRM-14B~\citep{khalifa2025processrewardmodelsthink} & 67.6  & 71.4  & 54.8  & 59.3  & 63.3  \\
        \noalign{\vskip 0.11cm}
        \hdashline
        \noalign{\vskip 0.11cm}
        \multicolumn{6}{l}{\textcolor{lightgray!99}{\textit{Our Critic}}} \\
        RefCritic-Qwen-14B & 81.9 & 71.2 & 58.1  & 60.7  & 68.0   \\
        RefCritic-R1-14B & \textbf{86.3}  & \textbf{82.0}  & \textbf{67.6}  & \textbf{72.3}  & \textbf{77.1}  \\
        \bottomrule
        \end{tabular}%
     }
    \caption{
        The evaluation results of PRMs, LLM as a critic, and RefCritic critic models on ProcessBench. The metric is the F1 score, the harmonic mean of precision for correct and incorrect solutions. All our critic models are followed by an extract model (Qwen2.5-14B-Instruct) to get the error step for easy evaluation. Content marked with "*" sourced from Processbench. As shown in Table \ref{tab:temp_process}, we use the same template as used in Processbench.
    }
        \label{tab:critic_performance}

\vspace{-5pt}
\end{table*}

\subsection{Main Results}
As shown in Table~\ref{tab:main_results}, we present the performance of RefCritic against various baselines on AIME24, AIME25, and OlympiadBench datasets.
In the one-round critique and refinement settings, RefCritic consistently provides the most effective feedback for policy model improvement, demonstrating the effectiveness of incorporating refinement performance as a reward in our reinforcement learning approach. Specifically, on the challenging AIME25 dataset, RefCritic-Qwen-14B and RefCritic-R1-14B enhance the policy model's $Pass@1$ performance by 6.8\% and 7.2\%, respectively, significantly outperforming both self-critique baselines and models trained via supervised fine-tuning. Similar patterns emerge across AIME24 and Olympiad benchmarks, confirming that directly optimizing for policy model refinement performance during RL training enables critic models to generate more actionable feedback.

When scaling up the policy model's response generation and applying critic model filtering, RefCritic achieves superior performance across nearly all experimental settings. For instance, on AIME25, RefCritic-RL improves $Maj_c@64$ with an average benefit of 3.6\%(1.1\% for RefCritic-Qwen and 6.1\% for RefCritic-R1). These results demonstrate that our refinement-oriented critic not only enhances feedback quality but also improves critical evaluation capabilities. Notably, as the sampling scale increases from 8 to 64, the overall performance gains from RefCritic become more pronounced, indicating our critic models' high discriminative accuracy in identifying and preserving high-quality solutions from larger candidate pools.

\subsection{Out-of-Distribution Performance}
\begin{table}[!t]
    \centering
    \resizebox{\linewidth}{!}{
        \begin{tabular}{l|c|cc}
        \toprule
        \multirow{2}[1]{*}{Model} & LiveCodeBench & \multicolumn{2}{c}{GPQA} \\
        \cmidrule(lr){2-3} \cmidrule(lr){3-4} & $Pass_r@1$ & $Pass_r@1$ & $Maj_c@16$ \\
            \midrule
            Qwen-14B Majority Vote&   18.9   & 19.5  & 23.3  \\
            Qwen-14B as Critic&   20.9    & 19.5  & 22.7  \\
            \noalign{\vskip 0.11cm}
            \hdashline
            \noalign{\vskip 0.11cm}
            \multicolumn{4}{l}{\textcolor{black}{\textit{RefCritic-R1-14B(Ours)}}} \\
            \quad $SFT$ &  21.5  & 19.2  & 22.8  \\
            \quad $RL_{\lambda=0}$   &  21.8   & 18.9  & 24.0  \\
            \quad $RL_{\lambda=0}\xrightarrow{after}RL_{\lambda=1}$    &  \textbf{22.9}   & \textbf{20.0} & \textbf{24.3} \\
            \midrule
            R1-Qwen-14B Majority Vote & 51.0  & 58.7  & 61.6  \\
            R1-Qwen-14B as Critic & 52.4  & 57.7  & 60.6  \\
            \noalign{\vskip 0.11cm}
            \hdashline
            \noalign{\vskip 0.11cm}
            \multicolumn{4}{l} 
            {\textcolor{black}{\textit{RefCritic-Qwen-14B(Ours)}}} \\
            \quad $SFT$ & 52.3  & 58.0  & 62.5  \\
            \quad $RL_{\lambda=0}$   & 53.6  & 59.0  & 64.6  \\
            \quad $RL_{\lambda=0}\xrightarrow{after}RL_{\lambda=1}$ & \textbf{54.1}  & \textbf{59.3} & \textbf{65.1} \\
            \bottomrule
        \end{tabular}
   }
    \caption{Performance comparison of different approaches on LiveCodeBench and GPQA.}
    \label{tab:ood_result}
\end{table}
Additionally, we also evaluated RefCritic on out-of-distribution tasks to demonstrate its generalizability. Considering that the model was trained on mathematical data, we chose to use LiveCodeBench to verify its performance on coding, and GPQA to evaluate its performance in challenging knowledge reasoning. We found that RefCritic still performs well on out-of-distribution benchmarks. Although the improvements are not as substantial as in the mathematical tasks, they still bring considerable gains. Specifically, RefCritic-R1-14B achieved a 3.1\% performance improvement on LCB\footnote{Since coding tasks cannot perform Majority Vote, we only report $Pass_r@1$ performance.}, and improved $Maj_c@64$ from 61.6\% to 65.1\% on the GPQA task, representing a 3.5\% performance gain. Similar progress also appeared in RefCritic-Qwen-14B. These results suggest that RefCritic's critic capabilities can be applied to a wide range of tasks.

\subsection{Critic Performance}
In this section, we evaluate RefCritic on ProcessBench to explore whether it can accurately identify true error locations in solutions. The experimental results presented in Table~\ref{tab:critic_performance} demonstrate that RefCritic significantly outperforms most previous baselines, including methods that utilize step-level supervision. RefCritic-Qwen achieves an average performance of 68, while RefCritic-R1 reaches an impressive 77 average performance. This indicates that our dual reward mechanism effectively guides the model in developing accurate error identification capabilities.

This finding is consistent with the growth of the model output length during RL training. Specifically, the average output length of RefCritic-Qwen increased from about 500 tokens to 3500 tokens, while RefCritic-R1 increased from 3000 tokens to 8000 tokens. This indicates the increasingly detailed critiques, making step-level critique possible.

These findings demonstrate that even without explicit step-level supervision, our approach enables critic models to develop a nuanced understanding of solution processes and identify errors with high precision. This capability is crucial for generating actionable feedback that can effectively guide policy models toward improved solutions.

\begin{figure*}[ht!]
    \centering
    \subfigure[Qwen2.5-72B Performance]{
        \includegraphics[width=0.31\textwidth]{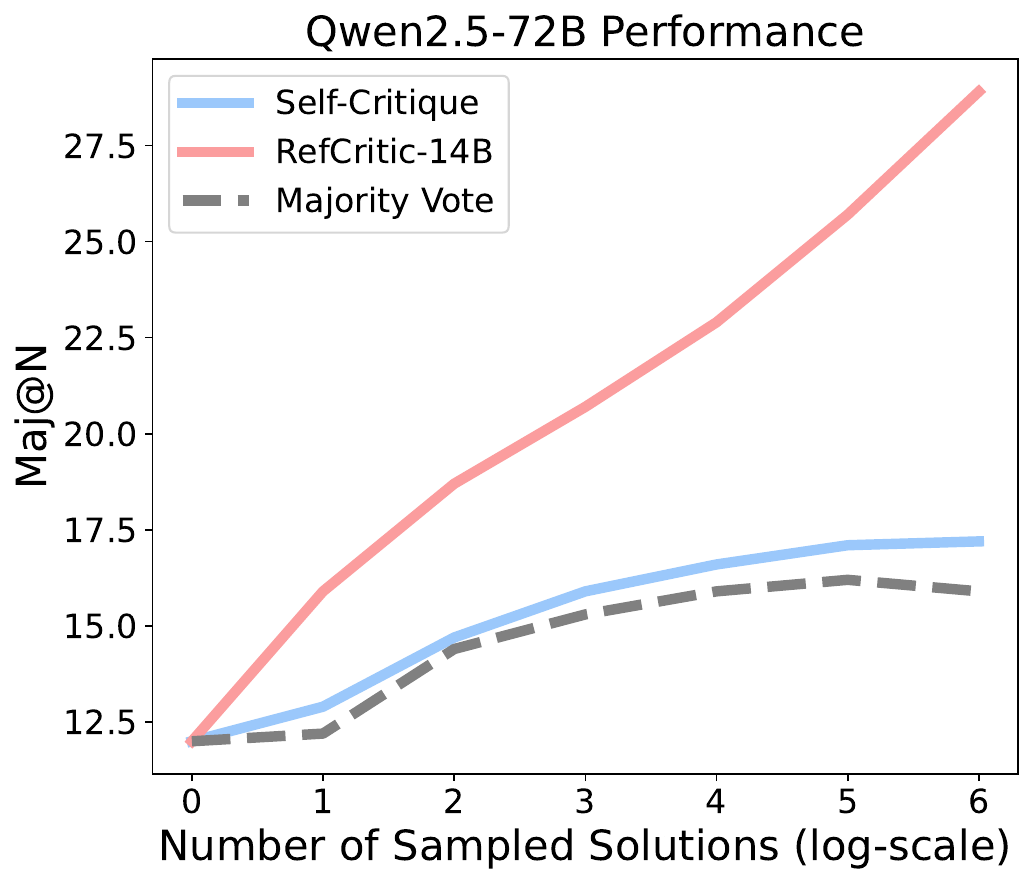}
    }
    \subfigure[DeepSeek-Distill-Qwen-32B Performance]
    {
        \includegraphics[width=0.31\textwidth]{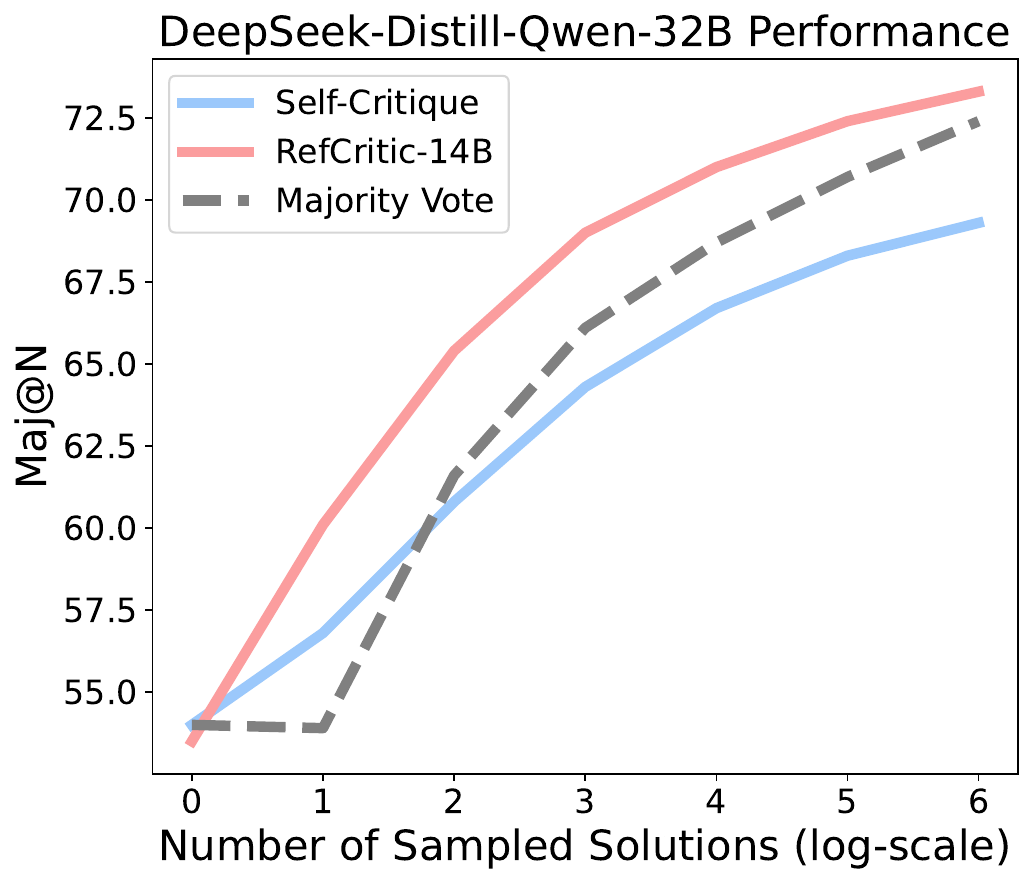}
    } 
    \subfigure[QwQ-32B Performance]
    {
        \includegraphics[width=0.31\textwidth]{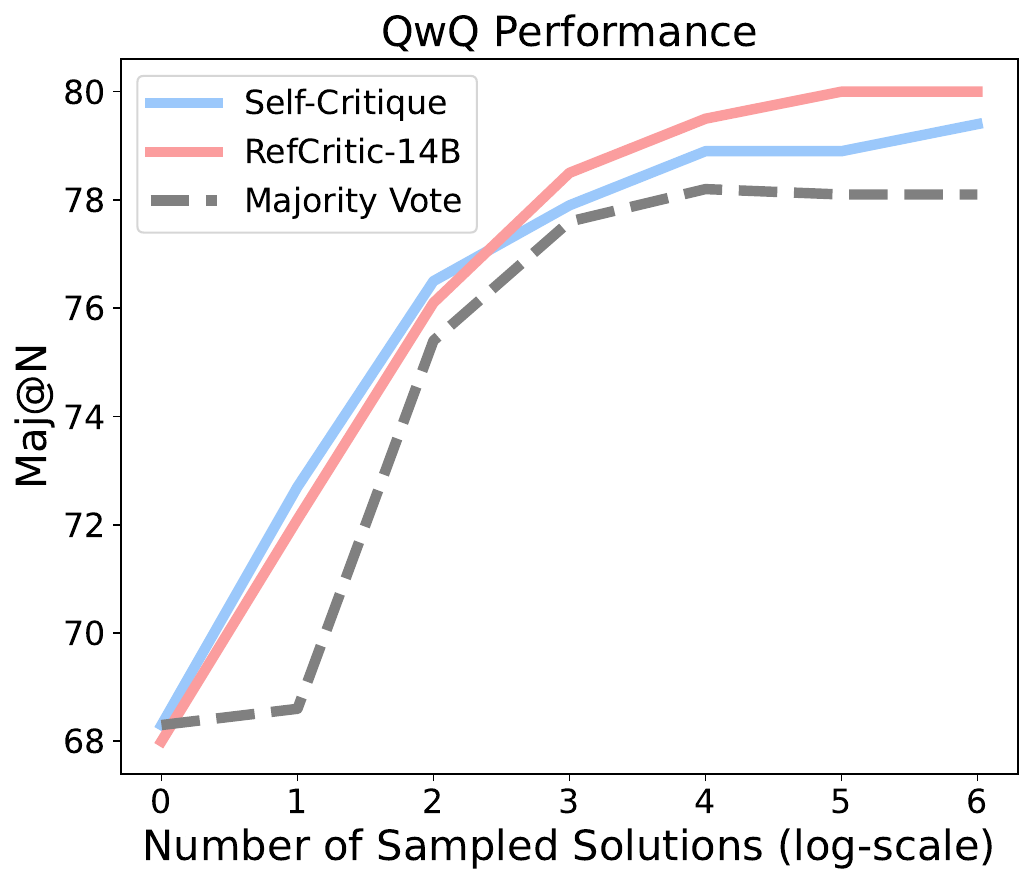}
    } 
    \caption{
        Supervision of RefCritic-R1-14B on stronger models like Qwen2.5-72B, DeepSeek-Distill-Qwen-32B, and QwQ-32B.
    }
    \label{fig:stronger}
\end{figure*}

\begin{figure*}[ht!]
    \centering
    \subfigure[Critique Scaling on RefCritic.]{
        \includegraphics[width=0.31\textwidth]{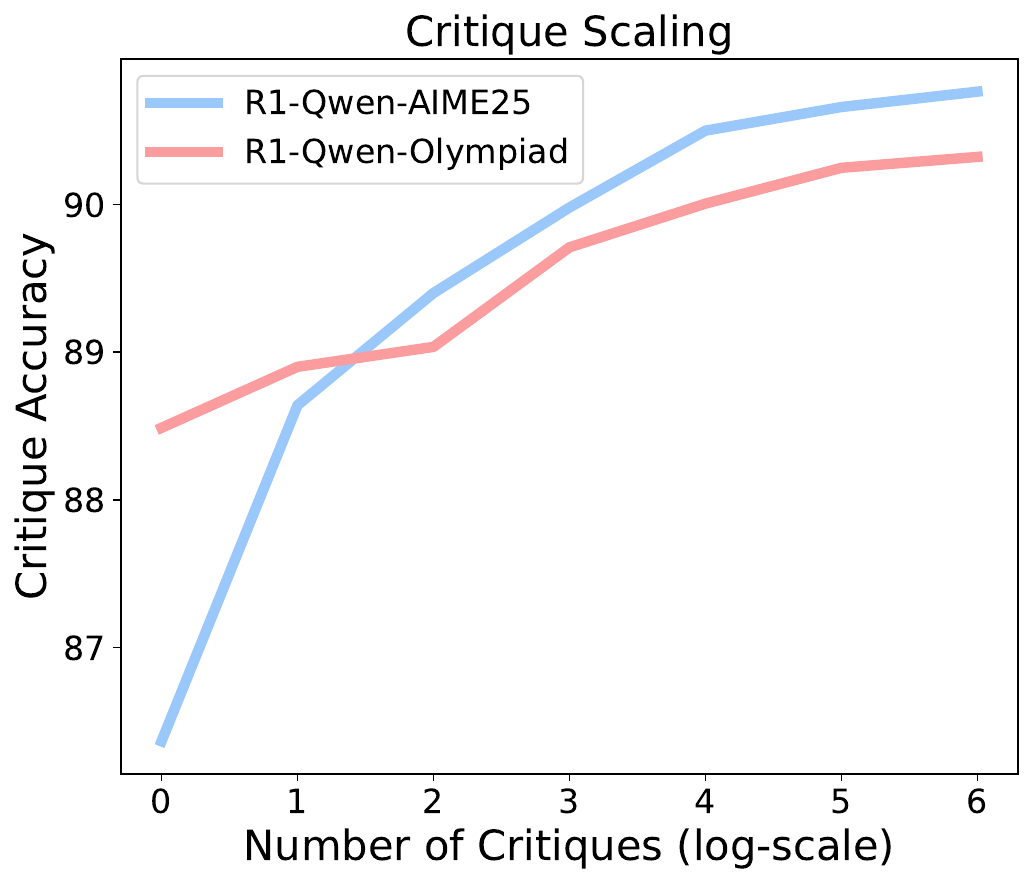}
    }
    \subfigure[Scaling on RefCritic-Qwen.]
    {
        \includegraphics[width=0.31\textwidth]{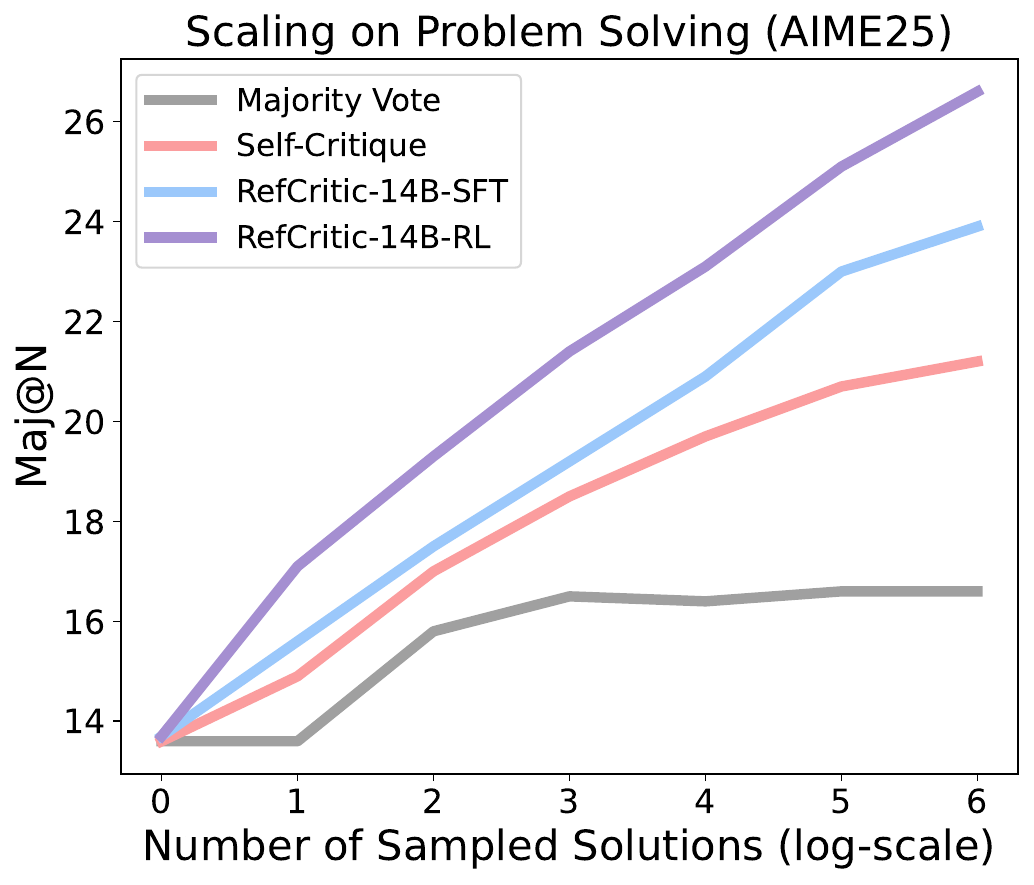}
    } 
    \subfigure[Scaling on RefCritic-R1.]
    {
        \includegraphics[width=0.31\textwidth]{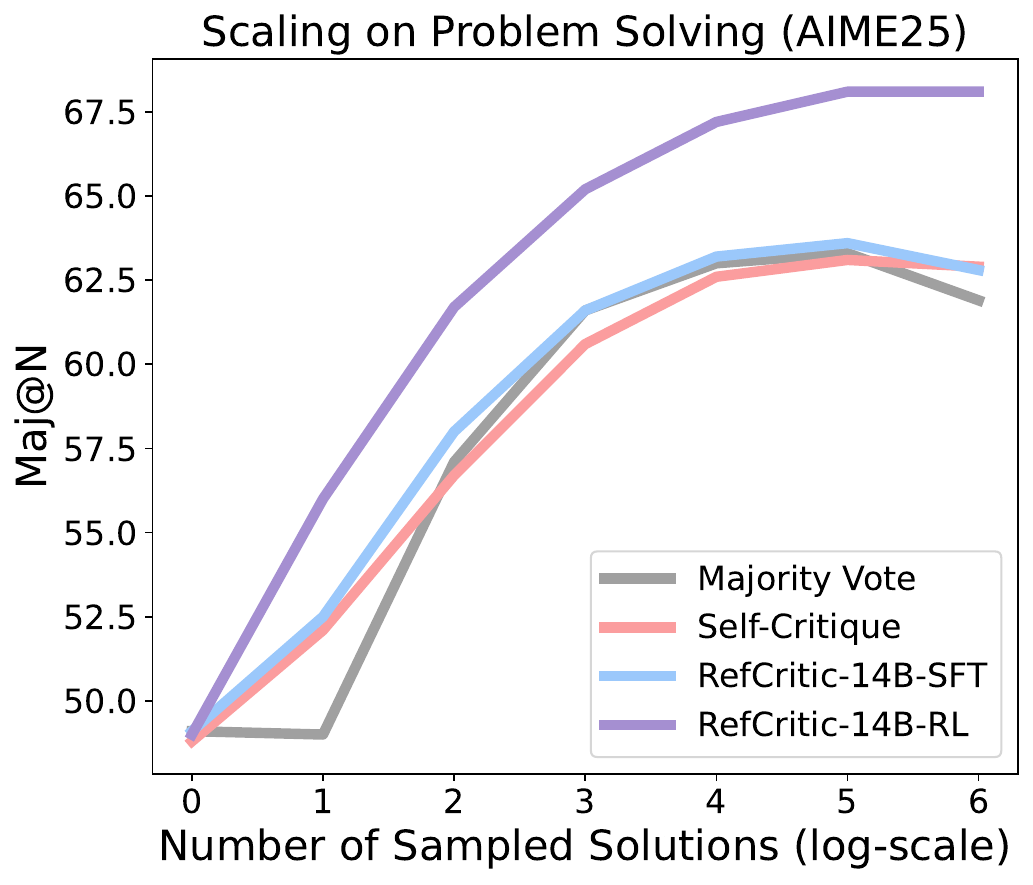}
    } 
    \caption{
        Test-time scaling with RefCritic. Subplot (a), scaling the number of critiques with RefCritic. Subplot (b), scaling the number of sampled solutions with RefCritic-Qwen-14B on AIME24. Subplot(c), scaling the number of sampled solutions with RefCritic-R1-14B on AIME25.
    }
    \label{fig:scaling_results}
\end{figure*}

\subsection{Test-time Scaling}

\paragraph{Scaling on Critique}
In this section, we investigate whether RefCritic's critique capabilities can benefit from test-time scaling. We construct two evaluation sets by sampling 16 solutions for each problem from AIME25 and Olympiad. For each question, we sample 64 critiques from RefCritic to examine whether increasing the number of critiques can progressively enhance critique performance.
The results presented in Figure~\ref{fig:scaling_results} (a) demonstrate that RefCritic's critique performance exhibits consistent improvement with increased sampling. On AIME25, performance steadily improves, reaching a 4\% increase as the number of sampled critiques increases. This finding suggests that RefCritic benefits from test-time scaling when multiple critiques are aggregated, thereby further enhancing its ability to evaluate mathematical solutions accurately. The scaling curve on Olympiad is relatively flat in comparison, which may be due to the difficulty of the tasks for the policy model (but a clear scaling trend can still be observed).

\paragraph{Scaling on Problem Solving}
Figure~\ref{fig:scaling_results} (b) and (c) illustrate the scaling performance of RefCritic-Qwen-14B and RefCritic-R1-14B on AIME25 as measured by maj@N with increasing rollout samples. The results demonstrate that RefCritic models consistently outperform baselines across different sampling scales, with the performance gap becoming more pronounced as the number of samples increases. This trend further validates that our refinement-oriented critic approach maintains its effectiveness advantage even in larger-scale inference scenarios.

\subsection{Supervision of Stronger Models}
In this section, we investigate whether RefCritic can effectively provide cross-model supervision for even more powerful reasoning models. We evaluate our approach using state-of-the-art models such as QwQ, DeepSeek-Distill-Qwen-32B, and Qwen 2.5-72B on the challenging AIME25 dataset. We compare three settings: (1) standard majority voting with the base models, (2) self-critique where the models evaluate their own solutions, and (3) cross-model supervision using our RefCritic-14B as the critic model.

As shown in Figure~\ref{fig:stronger}, even the most powerful reasoning and instruct models exhibit minimal or sometimes negative performance gains when employing self-critique compared to standard majority voting. This observation suggests a consistent limitation in these models' ability to critically evaluate their own solutions, regardless of their scale or overall reasoning capabilities. In contrast, our RefCritic approach consistently demonstrates positive improvements across nearly all experimental settings, even when supervising significantly larger and more capable reasoning models. Specifically, with 32 samples, RefCritic supervision improves QwQ performance by 1.5\% compared to standard majority voting and by 1.1\% compared to QwQ's self-critique approach. Similar patterns emerge for DeepSeek-Distill-Qwen-32B and Qwen 2.5-72B, confirming that RefCritic's benefits extend across model families and scales.

\subsection{Ablation} 
\begin{table}[!t]
    \centering
    \resizebox{\linewidth}{!}{
        \begin{tabular}{l|ccc}
        \toprule
        \multirow{2}[1]{*}{Model} & \multicolumn{1}{c}{AIME25} & \multicolumn{1}{c}{AIME24} & \multicolumn{1}{c}{Olympiad} \\
        \cmidrule(lr){2-2} \cmidrule(lr){3-3} \cmidrule(lr){4-4}
              & $Pass_r@1$ & $Pass_r@1$ & $Pass_r@1$ \\
            \midrule
            Qwen-14B as Critic & 14.5 & 13.7 & 45.8  \\
            \multicolumn{4}{l}{\textcolor{black}{\textit{RefCritic-Qwen-14B(Ours)}}} \\
            \quad $SFT$ & 15.0  & 15.2 & 46.6\\
            \quad $RL_{\lambda=0}$  & 18.5  & 19.1 & 51.4 \\
            \quad $RL_{\lambda=1}$  & 19.5 & 21.4 & 54.3 \\
            \quad $RL_{\lambda=0} \xrightarrow{after} RL_{\lambda=0}$ & 19.6 & 21.7 & 53.6 \\
            \quad $RL_{\lambda=0} \xrightarrow{after} RL_{\lambda=1}$ & \textbf{21.2} & \textbf{23.0} & \textbf{55.7} \\
            \bottomrule
        \end{tabular}
        }
    \caption{Ablation results on RefCritic-Qwen-14B.}
    \label{tab:ablation}
\end{table}

Finally, we also propose some ablation studies to understand the role of the two RL training stages in RefCritic, namely $\lambda$=0 and $\lambda$=1. Considering the training cost, we only conduct experiments on RefCritic-Qwen-14B, and ablations on DeepSeek-Distill-Qwen-14B will be added in future research. Specifically, we aim to explore the importance of Refinement Reward. To this end, we mainly compared two groups of experiments: 1) $RL_{\lambda=0}$ and $RL_{\lambda=1}$. 2) $RL_{\lambda=0}$ $\xrightarrow{after}$ $RL_{\lambda=0}$ and $RL_{\lambda=0}$ $\xrightarrow{after}$ $RL_{\lambda=1}$.
Each group of experiments is optimized with the same parameters. The results of all these ablation experiments are shown in Table~\ref{tab:ablation}. As expected, under the same settings, refinement reward improves the refinement performance of models. Furthermore, first using $RL_{\lambda=0}$ for Critic optimization is also beneficial to RefCritic. $RL_{\lambda=0}$ can quickly improve Critic performance at a lower cost, making $RL_{\lambda=0}$ $\xrightarrow{after}$ $RL_{\lambda=1}$ a setting that balances cost and performance.
\section{Conclusion}
In this work, we introduced RefCritic, a novel approach for training critic models to critique the correctness of solutions and provide effective refinement feedbacks from LLMs.
Our method leverages a dual-reward system that jointly optimizes for judgment accuracy and refinement effectiveness, creating an explicit feedback loop between critique quality and policy model improvement.
Our experiments demonstrated that while SFT alone is insufficient for producing comprehensive critiques despite generating better critiques, the integration of reinforcement learning with our designed reward signals significantly enhances both the analytical depth and practical utility of critiques.
Experimental results across challenging mathematical datasets and out-of-distribution benchmarks validate RefCritic's effectiveness in consistently enhancing policy model performance in both critique-refinement and majority vote settings.
Further experiments on ProcessBench demonstrate that even without a step-level signal, RefCritic can effectively identify the error step.

\section*{Limitations}

Despite the promising results, RefCritic has several limitations. The dual-reward reinforcement learning framework requires significant computational resources, which may limit its scalability for very large models. Our approach primarily focuses on mathematical and logical reasoning tasks, and its generalizability to domains like commonsense reasoning or specialized professional contexts remains to be thoroughly investigated. 

\bibliography{custom}

\appendix
\onecolumn
\newpage
\section{Data Construction}
\label{data_filter}
We filtered about 120k problems from the 900k mathematical problems of NuminaMath-1.5 \cite{numina_math_datasets}. Detailed filter process and utilization can be found in the section \ref{data_filter}.
Our training data pipeline involves rigorous filtering to ensure high-quality and diverse mathematical problems.
\paragraph{Problem Deduplication} We start with a deduplication process on the 900k mathematical problems from NuminaMath-1.5 \cite{numina_math_datasets}. The deduplication process includes the string-based process by performing exact matching after removing special characters such as '\$', '[', ']', etc., and semantic deduplication, where we used gte-multilingual-base embeddings to compute cosine similarity between problem pairs and removing those with similarity scores exceeding 0.95.
\paragraph{Problem Filter} Then, we utilize Qwen2.5-72B-Instruct as a judge to filter problems based on several criteria: unsolvable problems, proof problems requiring formal mathematical proofs, and multiple-choice problems. We sample eight solutions for each problem with DeepSeek-Distill-Qwen-7B to ensure appropriate difficulty distribution. We remove problems where DeepSeek-Distill-Qwen-7B either solves all eight attempts correctly or fails on all eight attempts, thus eliminating trivial or impossibly difficult issues. After this comprehensive filtering process, we obtain approximately 120k high-quality mathematical problems for training. All prompt templates are provided in the Appendix.
\paragraph{Solution Sampling} To create training data for critic models, we sample 8 responses from the policy model for each problem, remove problems where all solutions are correct or incorrect, filter out incomplete generations, and ensure balanced training by retaining at most two responses per problem (one correct and one incorrect). For efficient scaling, all responses are sampled by sglang inference services\footnote{https://github.com/sgl-project/sglang}.

\section{Templates}
\label{template}
\begin{table*}[h]
\centering
\small
\begin{minipage}{0.9\textwidth}
\begin{mdframed}
\begin{verbatim}
Given a student's mathematical solution, analyze it step-by-step to determine correctness. 
Do not solve the problem yourself, provide feedback focus on the student's work to help 
them learn. Conclude your feedback as:

**Correctness**: Correct | Incorrect
(If incorrect)
**Comment**: Identify the specific error in the solution and help the student recognize 
why their approach leads to an incorrect result. Then, provide a comment that will help 
the student to resolve this problem.

Do not expose any answer!

[Problem]
{problem}

[Solution]
{solution}
\end{verbatim}
\end{mdframed}
\end{minipage}
\caption{The template we used for critique.}
\label{tab:temp_critic}
\end{table*}
\begin{table*}[ht]
\centering
\small
\begin{minipage}{0.9\textwidth}
\begin{mdframed}
\begin{verbatim}
Review your solution to a mathematical problem and a feedback from your teacher. Create an 
improved version that fixes the identified errors.
Please reason step by step, and put your final answer within \\boxed{{}}.

[Problem]
{problem}

[Original Solution]
{solution}

[Teacher Feedback]
{critique}
\end{verbatim}
\end{mdframed}
\end{minipage}
\caption{The template we used for refinement.}
\label{tab:temp_refine}
\end{table*}
\begin{table*}[ht]
\centering
\small
\begin{minipage}{0.9\textwidth}
\begin{mdframed}
\begin{verbatim}
The following is a math problem and a solution (split into paragraphs, enclosed with tags and
indexed from 0):

[Math Problem]

{problem}

[Solution]

{solution}

Your task is to review and critique the solution paragraph by paragraph. Once you identify an 
error in a paragraph, return the index of the paragraph where the earliest error occurs. 
Otherwise, return the index of -1 (which typically denotes "not found").

Please put your final answer (i.e., the index) in \boxed{}.
\end{verbatim}
\end{mdframed}
\end{minipage}
\caption{The template we used for evaluating processbench.}
\label{tab:temp_process}
\end{table*}
\label{sec:appendix}
\end{document}